\documentclass[conference]{IEEEtran}
\IEEEoverridecommandlockouts
\usepackage{colortbl} 
\usepackage{ragged2e} 
\usepackage{caption}  
\usepackage{float}    
\usepackage{cite}
\usepackage{amsmath,amssymb,amsfonts}
\usepackage{algorithmic}
\usepackage{graphicx}
\usepackage{textcomp}
\usepackage{enumitem}

\usepackage{xcolor}
\usepackage{longtable}
\usepackage{array}
\usepackage{graphicx} 
\usepackage{caption}  
\usepackage{float}    
\usepackage{hyperref} 
\usepackage{url}      
 \usepackage{booktabs}
 \def\BibTeX{{\rm B\kern-.05em{\sc i\kern-.025em b}\kern-.08em
     T\kern-.1667em\lower.7ex\hbox{E}\kern-.125emX}}
    \usepackage{caption}
\begin{document}

\title{Unified Large Language Models for Misinformation Detection in Low-Resource Linguistic Settings\\
{\footnotesize \textsuperscript{}}

}

\author{
\IEEEauthorblockN{Muhammad Islam}
\IEEEauthorblockA{\textit{College of Science and Engineering} \\
\textit{James Cook University} \\
Cairns, Australia \\
muhammad.islam1@my.jcu.edu.au}
\and
\IEEEauthorblockN{Javed Ali Khan}
\IEEEauthorblockA{\textit{Department of Computer Science} \\
\textit{University of Hertfordshire} \\
Hatfield, United Kingdom of Great Britain and Northern Ireland \\
J.a.khan@herts.ac.uk}
\and
\IEEEauthorblockN{Mohammed Abaker}
\IEEEauthorblockA{\textit{Department of Computer Science} \\
\textit{Applied College, King Khalid University} \\
Muhayil 61913, Saudi Arabia \\
moadam@kku.edu.sa}
\and
\IEEEauthorblockN{Ali Daud}
\IEEEauthorblockA{\textit{Faculty of Resilience} \\
\textit{Rabdan Academy, Abu Dhabi} \\
United Arab Emirates \\
alimsdb@gmail.com}
\and
\IEEEauthorblockN{Azeem Irshad{*}\thanks{*Corresponding author: azeemirshad@gpgcam.edu.pk (Azeem Irshad)}}
\IEEEauthorblockA{\textit{Faculty of Computer Science} \\
\textit{GGC Asghar Mall, Rawalpindi, Punjab HED} \\
Pakistan \\
azeemirshad@gpgcam.edu.pk}
}

\maketitle

\begin{abstract}
The rapid expansion of social media platforms has significantly increased the dissemination of forged content and misinformation, making the detection of fake news a critical area of research. Although fact-checking efforts predominantly focus on English-language news, there is a noticeable gap in resources and strategies to detect news in regional languages, such as Urdu. Advanced Fake News Detection (FND) techniques rely heavily on large, accurately labeled datasets. However, FND in under-resourced languages like Urdu faces substantial challenges due to the scarcity of extensive corpora and the lack of validated lexical resources. Current Urdu fake news datasets are often domain-specific and inaccessible to the public. They also lack human verification, relying mainly on unverified English-to-Urdu translations, which compromises their reliability in practical applications. This study highlights the necessity of developing reliable, expert-verified, and domain-independent Urdu-enhanced FND datasets to improve fake news detection in Urdu and other resource-constrained languages. This paper presents the first benchmark large FND dataset for Urdu news, which is publicly available for validation and deep analysis. We also evaluate this dataset using multiple state-of-the-art pre-trained large language models (LLMs), such as XLNet, mBERT, XLM-RoBERTa, RoBERTa, DistilBERT, and DeBERTa. Additionally, we propose a unified LLM model that outperforms the others with different embedding and feature extraction techniques. The performance of these models is compared based on accuracy, F1 score, precision, recall, and human judgment for vetting the sample results of news. The proposed model outperforms advanced machine learning and deep learning models previously used in the literature for fake news detection. The dataset and code for the experiments are publicly available at the GitHub link: \url{https://github.com/MislamSatti/Urdu-Large-language-dataset}.
\end{abstract}

\begin{IEEEkeywords}
Unified framework , multilingual, low resource,  LLM, Pre-Trained Model, multi-head attention, embedding, NLP, ML, DL, Stacking 
\end{IEEEkeywords}

\section{Introduction}

Over the past few decades, the internet has become a part of our daily lives, making information more accessible than ever before. As a result, traditional print media has taken a backseat to online news platforms, where news spreads rapidly. However, not all news is reliable and legit. Fake News (FN), which can take various forms, such as partial truth, deception, or outright fabrication, has become a significant concern in the digital age \cite{1},\cite{2}.

The spread of FN has become a pressing issue in today's digital world \cite{2a}, especially during events like the COVID-19 pandemic. False information about COVID-19 \cite{3}, such as misleading claims about its origin, supposed remedies like gargling bleach, or unfounded links to 5G technology, has led to confusion and even harm. Similarly, during conflicts like the airstrikes between two neighboring countries, India and Pakistan, or the war between Russia and Ukraine, FN has flooded online platforms like social network websites(SNWs), exacerbating tensions and spreading misinformation\cite{4}. Therefore, these instances underscore the urgent and dire need for Fake News Detection (FND) to mitigate the societal risks of deceptive information dissemination.

Detecting fake news often relies on human judgment, but this approach isn't always dependable. It requires a deep understanding of the subject matter to discern fake news accurately\cite{5}. Hence, there's a growing need for automated methods to detect fake news more effectively.

Having a high-quality dataset is crucial for automatically detecting fake news using artificial intelligence techniques. The dataset is the foundation for training and evaluating the model’s performance. Due to abundant resources and large datasets, researchers have extensively studied fake news detection in languages like English. Well-known datasets like LIAR \cite{6}, fake or real news \cite{7}, Twitter News, and  Weibo News \cite{8} have been established. While these datasets encompass news across various subjects, none of them are tailored to the news of specific regions or countries or in different languages.

Several fact-checking platforms, like PolitiFact.com, FactCheck.org, and Snopes.com, maintain databases of verified information for non-fictional news articles in English. These websites leverage digital libraries and advanced technology to assess the truthfulness and credibility of factual claims within the content but not for low-resource languages like Urdu. As a result, these fact-checked assertions cannot rectify the inaccurate understandings that users may have about events in Urdu.

A low-resource language like Urdu is among the top ten most spoken languages, with around 230M people speaking  worldwide\cite{4, 4a}. The absence of suitable language processing tools and standardized datasets adds to the challenge of classifying fake news in Urdu. The exponential growth of Urdu news spreads electronically daily, and a mechanism needs to be established to check its legitimacy. The growth of literature \cite{2}, \cite{3}, \cite{9}, \cite{10}in FN detection determines its importance as well, but mostly, the datasets are smaller in size\cite{9}, domain-constrained, publicly unavailable, and are in Roman Urdu and not authentic sources. Therefore, there is a dire need for a state-of-the-art (SOTA) large benchmark corpus not translated from English to Urdu and in multi-domain to need an automatic technique to detect fake news. This research curated the first Large Urdu Language dataset (LUND) of 27.5K samples for Legitimate and fake news with total words of 106,221K and 84.07K unique words.  

The other foremost concern is the availability of automation techniques to validate the legitimacy of the news shared in regional languages on the web portals (WP) and SNWs. Numerous machine learning (ML), deep learning (DL), and natural language processing (NLP) algorithms with different embedding are helpful in evaluating the correctness of the datasets, \cite{11}. NLP is used in text processing tasks, mainly generation and classification, fake or legit news detection, and named entity recognition (NER). It is also very helpful in text classification, identifying word pattern sentiment analysis, and term frequency\cite{12}.

The performance of advanced large language models such as BERT, GPT2, XLNet, and DeBerta has been analyzed in previous studies \cite{2,4,9,9a,9b}, with comparisons based on various metrics including accuracy, precision, recall, and F1 score. These models were evaluated on smaller individual datasets. In contrast, our work jointly trains on these existing datasets and proposes a unified dataset for cross-dataset evaluation to assess their generalization ability and robustness. Therefore, this research applies recent pre-trained NLP large language models \cite{3}, such as XLM-Roberta and mBERT, to a significant benchmark Urdu news dataset, which is commonly used for text understanding, classification and generation tasks. We compare these models with established state-of-the-art (SOTA) ML and DL algorithms \cite{1}, \cite{9}, \cite{13}, \cite{14}. Finally, we evaluate the performance of proposed model against other models using our proposed dataset, and misclassified samples are reviewed with expert journalists for human judgment.

\begin{figure}[thbp]
 \centering
 \includegraphics[width=0.5\textwidth,height=5cm]
 {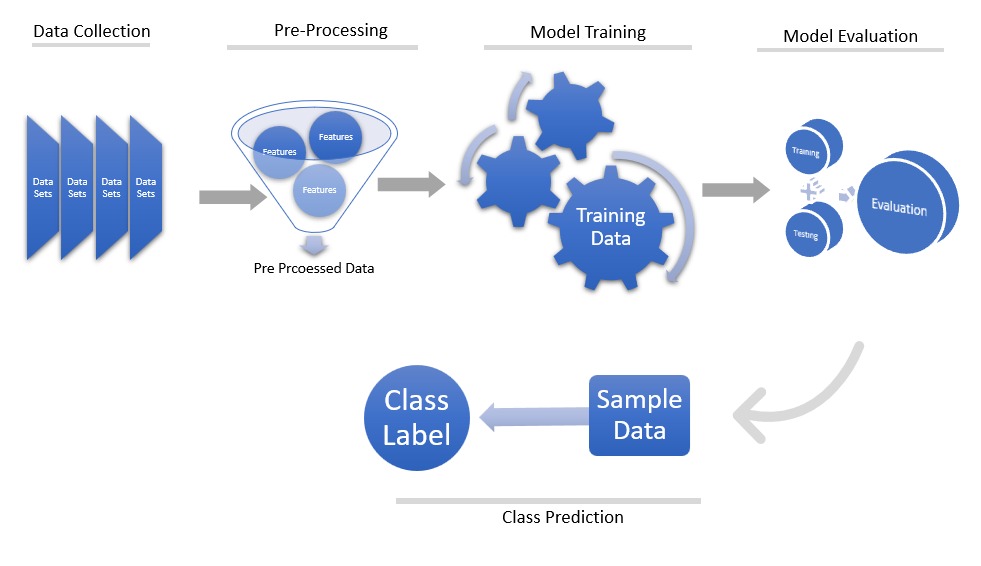}
\caption{The key Steps Performed in Methodology.}
\label{fig}
\end{figure}

\vspace{5pt}




%
\subsection{Contributions}
This research study endeavors to assess the reliability and generalization factor of Large Urdu news dataset, in general. Moreover, it attempts to determine the capability of Large language model to perform in detecting fake news for low resource language. Specifically, the following are the major contributions of the research.
\begin{itemize}
\item We tackle key challenges in fake news dataset preprocessing and develop a generalized benchmark dataset for Urdu fake news detection. Our contributions include resolving data inconsistencies, standardizing labeling schemes, addressing domain and class imbalances, and eliminating duplication. By ensuring lexical diversity and harmonizing data formats, we provide a structured, balanced, and reliable dataset, enhancing the reliability and generalizability of Urdu fake news detection models.
\item 	We conduct an in-depth analysis and experimentation of existing models for fake news detection in low-resource languages. We propose an unified framework with better attention to the critical words to evaluate the benchmark dataset and assess model performance with key evaluation metrics. Our study compares ML, DL, and pre-trained Large Language Models (LLMs) with transformer backbones, incorporating expert journalist evaluations to enhance reliability and effectiveness.
\end{itemize}
\subsection{Structural setup}
The paper is divided into different sections. The first section is the introduction. The second section reviews the existing literature. The third is related to preparing the dataset and experimenting with LLM on the benchmark dataset. The fourth section is based on the results and evaluation of the models and their comparison with other famous ML and DL models. Lastly, the last section concludes the literature and future work. Figure 1 demonstrates the abstract-level view of the proposed methodology.

\section{Related Work}

In recent years, Fake News Detection (FND) has garnered significant attention, with researchers exploring various methodologies to combat the proliferation of misinformation. Despite this surge in research, Urdu FND remains relatively unexplored. However, advancements in Machine Learning (ML), Deep Learning (DL), and Natural Language Processing (NLP) have paved the way for innovative approaches in Urdu FND. This study assesses the effectiveness of state-of-the-art (SOTA) techniques for Urdu FND and classification, considering the existing benchmark datasets and their associated limitations.

Detecting fake news in Arabic is a critical challenge due to dialectal variations, semantic ambiguity, and cultural context. Recent research by the Azzeh et al. \cite{15} addresses this issue by creating a large annotated corpus of Arabic fake news. They explore transformer-based models like CAMeLBERT, achieving accurate text representations. While deep learning  \cite{15a}classifiers outperform traditional methods, choosing the best model depends on evaluation metrics. This work contributes valuable insights to the field of Arabic fake news detection, but some of the limitations are like manual annotation is overhead, and this model needs to be applied to a large, valid, and accurate dataset.
The FIRE 2020 \cite{2} task on Urdu fake news detection employed a dataset with 900 training and 400 testing articles, with the BERT-based approach demonstrating superior performance. In contrast, the UrduFake 2021 \cite{14}  paper utilized an ensemble of machine-learning models, achieving a macro F1-score of 0.552 and an accuracy of 0.713. While both studies explored various techniques, including character bi-grams and TF-IDF, the absence of a benchmarked corpus remains challenging in Urdu fake news detection. Despite advancements, further investigation is warranted to address this gap and enhance the robustness of detection methods.

BiL-FaND \cite{16} demonstrates notable efficacy in discerning misinformation across English and Urdu. Leveraging a combination of Multilingual BERT, LSTM models, and multimedia analysis techniques, the system achieves a commendable overall accuracy rate of 92.07\% in identifying fabricated news content in bilingual contexts. Specifically engineered to address the challenge of counterfeit news detection in both English and Urdu, BiL-FaND is distinguished by its utilization of advanced machine-learning methodologies and innovative architectural frameworks.

Authors \cite{17} discuss the introduction of the CLE Meeting Corpus for Urdu meeting summarization, evaluating various deep learning models, and addressing challenges within diverse organizational contexts. The study employs models such as ur-mT5-small, ur-mT5-base, ur-mBART-large, ur-RoBERTa, urduhack-small, and GPT-3.5. Notably, the ur-mT5-base and ur-mBART-large models demonstrate superior performance post-fine-tuning on meeting data, while the review also acknowledges limitations such as domain specificity and coherence challenges.

This paper \cite{18} outlines a novel approach for detecting violence incitement within Urdu tweets, employing a 1D-CNN model. This model demonstrates a notable accuracy of 89.84\%, surpassing alternative machine learning methodologies. The exploration extends to semantic, word embedding, and language models to establish contextualized representations. Additionally, the review delves into treating violence incitement as a binary classification task and the comparative analysis of various machine learning and feature models. The proposed framework underscores identifying violence-inciting content within Urdu tweets., complemented by examining semantic and contextual embedding models tailored for the Urdu language.

The study \cite{19} focuses on detecting hate speech in Roman Urdu, a crucial aspect given the prevalence of such content on social media platforms. Utilizing the HS-RU-20 corpus, comprising 5,000 annotated tweets, the research employs six supervised learning models, including Logistic Regression and Convolutional Neural Network. Logistic Regression and Support Vector Machine exhibit high accuracy in distinguishing between neutral and hostile speech, achieving 81\%  and 87\% accuracy, respectively. However, a significant challenge lies in ensuring the scalability and adaptability of improved annotation quality within the dataset, underscoring the need for further research in this area.

Recent developments in sentiment analysis \cite{20} encompass transitioning from traditional lexicon-based methods to deep learning approaches, leveraging convolutional neural networks (CNNs) and character-level embeddings. These techniques exhibit promise in multilingual sentiment classification tasks, where models like mBERT and mT 5 trained on 79 datasets and 27 languages, one language performs effectively on translated reviews in various languages. 

Detecting sarcasm in social media texts poses a significant challenge due to its context-based nature, rendering rule-based approaches less effective. To address this, recent research \cite{21} has focused on employing deep-learning classifiers and transformer architectures for improved sarcasm detection. Specifically, in the context of Urdu tweets, a novel mBERT-BiLSTM-MHA hybrid model has been introduced to tackle this issue. 

The study \cite{12} on Urdu Named Entity Recognition (U-NER) utilizes BERT and ELMo embedding, fine-tuned with Urdu data, achieving a high F1-score of 93.99\%. Challenges stemming from Urdu's morphological diversity and limited training data are addressed through data augmentation techniques. Future research aims to explore alternative embedding methods such as Flair and GPT to extend the model's applicability to other low-resource South Asian languages.
\begin{figure*}[]
 \centering
 \includegraphics[width=0.4\textwidth,height=7cm]
 {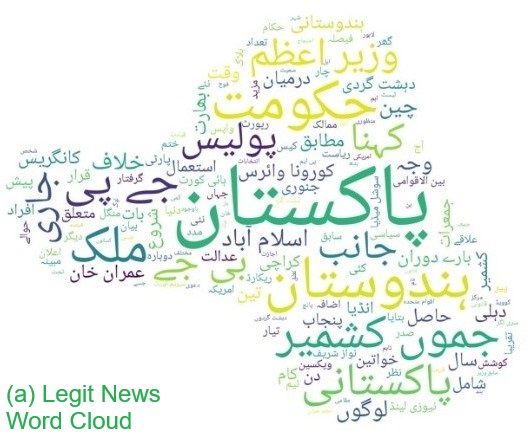}
 \includegraphics[width=0.4\textwidth,height=7cm]
 {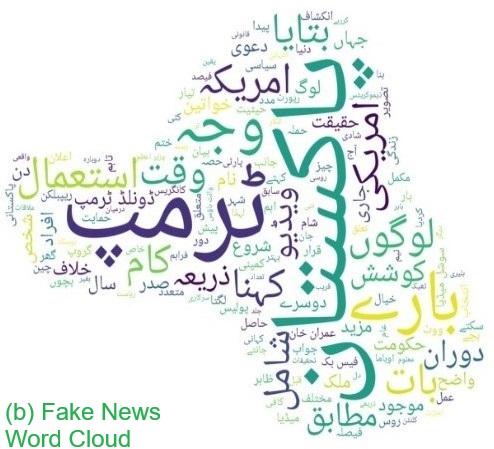}
\caption{word cloud of (a) legit (b) fake and  Urdu news.}
\label{fig}
\end{figure*}
A novel dataset \cite{1} for detecting fake news in Pakistan has been developed, marking a significant advancement in AI-driven misinformation detection. Through experimentation with various AI techniques and embeddings, LSTM with GloVe embeddings emerges as the top performer, achieving a notable F1-score of 0.94. Additionally, efforts to address dataset imbalances and potential sub-classifications underscore a commitment to robustness and future improvement in fake news detection methodologies tailored to the Pakistani context.

A recent study \cite{22}\cite{23} proposes a hybrid approach for detecting fake news in low-resource languages. This method combines extractive and abstractive summarization techniques, utilizing both YAKE and T5 models. It addresses challenges related to lengthy texts and multilingual input handling without translation. The study also compares mBERT Baseline, XLM-RoBERTa, and Semantic graph-based topic modeling, highlighting potential limitations of pre-trained language models (PLMs) in context preservation and mT5 performance in low-resource languages.
\arrayrulecolor{blue}
\renewcommand{\arraystretch}{2}
\begin{table*}[]
\captionsetup{labelfont={color=blue}, textfont={color=blue}} 
\caption{Comprehensive comparisons of models, dataset and evaluations results exhibited.}
\centering

\scriptsize
\begin{tabular}{p{0.8cm} p{3cm} p{0.5cm} p{2.5cm} p{2cm} p{1.3cm} p{2.5cm} p{2.5cm}}
\toprule
\textbf{\textcolor{blue}{Year}} & \textbf{Title} & \textbf{Venue} & \textbf{Research Work} & \textbf{Model Used} & \textbf{\textcolor{blue}{Language Support}} & \textbf{Results} & \textbf{Gaps / Future Work} \\
\midrule
\textcolor{blue}{2024} & Arabic Fake News Detection in Social Media Context Using Word Embeddings and Pre-trained Transformers & AJSE & Developed a corpus and applied LLM models & ARBERT, AraBERT, CAMeLBERT & \textcolor{blue}{Arabic} & CAMeLBERT achieved best results: Accuracy 72.6, Precision 70.6, Recall 72.0, F1 Score 71.3 & Manual annotation overhead; model needs a larger, more accurate dataset \\

\textcolor{blue}{2023} & Fake news detection on Pakistani news using machine learning and deep learning & ESA & Developed dataset for detecting fake news in Pakistan using AI techniques & LSTM with GloVe embeddings & \textcolor{blue}{English, Urdu} & LSTM with GloVe embeddings showed best performance with 0.94 F1-score & Dataset imbalance; potential sub-classification of labels \\

\textcolor{blue}{2024} & Enriching Urdu NER with BERT Embedding, Data Augmentation, and Hybrid Encoder-CNN Architecture & ACM TALLIP & Utilized BERT and ELMo embeddings fine-tuned with Urdu data & Word2vec, GloVe, FastText, BERT, and ELMo & \textcolor{blue}{Urdu} & Data augmentation attained highest F1-score of 93.99\% & Challenges due to morphological diversity and data scarcity; future work exploring Flair and GPT \\

\textcolor{blue}{2024} & Detection of Sarcasm in Urdu Tweets Using Deep Learning and Transformer-Based Hybrid Approaches & IEEE Access & Created Urdu sarcasm dataset (12,910 tweets) & CNN, LSTM, GRU, BiLSTM, CNN-LSTM, mBERT-BiLSTM-MHA & \textcolor{blue}{Urdu} & BiLSTM-MHA hybrid model accuracy of 79.51\%, F1-score of 80.04\% & Baseline model used for Urdu but needs application on Roman Urdu and other low-resource languages \\

\textcolor{blue}{2023} & Massively Multilingual Corpus of Sentiment Datasets and Multi-faceted Sentiment Classification Benchmark & NeurIPS & Constructed multilingual corpus of 79 datasets in 27 languages & mBERT, XLM-R, mT5 & \textcolor{blue}{27 languages} & mBERT performed well & Bias in Indo-European languages, dataset quality issues, lack of annotation coherence \\

\textcolor{blue}{2024} & Hate Speech Detection in Roman Urdu using Machine Learning Techniques & ICACS & Utilized HS-RU-20 corpus (5,000 Roman Urdu tweets) & Logistic Regression, Multinomial Naïve Bayes, CNN & \textcolor{blue}{Roman Urdu} & Logistic Regression and SVM achieved high accuracy (0.81 and 0.87 respectively) & Challenges in scalability and annotation quality \\

\textcolor{blue}{2024} & Detection of violence incitation expressions in Urdu tweets using convolutional neural network & ESA & Proposed framework for detecting violence incitation content & Urdu-RoBERTa, Urdu-BERT, 1D CNN & \textcolor{blue}{English, Urdu} & 1D CNN model outperformed with 89.84\% accuracy, 89.80\% macro F1-score & Limited generalizability, dataset size constraints, specific to Nastaliq Urdu \\

\textcolor{blue}{2024} & Meeting the challenge: A benchmark corpus for automated Urdu meeting summarization & IPM & Introduced CLE Meeting Corpus for Urdu meeting summarization & ur\_mT5, ur\_mBART, GPT-3.5 & \textcolor{blue}{Urdu, Hindi} & ur\_mT5-base and ur\_mBART-large outperformed GPT-3.5 & Domain specificity, lack of action item annotation, coherence challenges \\

\textcolor{blue}{2024} & Ax-to-Grind Urdu: Benchmark Dataset for Urdu Fake News Detection & TrustCom & Introduced a benchmark fake news dataset for Urdu & mBERT, XLNet, XLM-RoBERTa & \textcolor{blue}{Urdu, Hindi} & Ensemble model: Accuracy 0.9, Precision 0.91, Recall 0.89, F1-score 0.9 & Needs large dataset across domains, multimodal feature exploration \\

\textcolor{blue}{2022} & On the transferability of pre-trained language models for low-resource programming languages & ACM ICPC & Studied PLMs for Code Summarization and Code Search & Monolingual vs. multilingual PLMs & \textcolor{blue}{English} & Monolingual PLMs have better Performance-to-Time Ratio & Need for better dataset discussion and evaluation metrics \\

\textcolor{blue}{2024} & ChatGPT vs Gemini vs LLaMA on Multilingual Sentiment Analysis & arXiv & Evaluated LLMs on multilingual sentiment analysis across 10 languages & ChatGPT, Gemini, LLaMA2 & \textcolor{blue}{10 under-resourced languages} & ChatGPT and Gemini performed well; LLaMA2 showed optimistic bias & Scarcity of labeled data, challenges in sarcasm detection, ethical concerns \\

\textcolor{blue}{2024} & Fake news detection in low-resource languages: A novel hybrid summarization approach & KBS & Proposed transformer-based framework with hybrid summarization & mBERT, XLM-RoBERTa, Semantic graph-based topic modeling & \textcolor{blue}{Multilingual} & Best performance on Indonesian language (F1-score: 90.55\%) & PLMs struggle with long text processing \\

\textcolor{blue}{2024} & RumorLLM: A Rumor Large Language Model-Based Fake News Detection Data-Augmentation Approach & App. Sciences & Proposes Rumor Large Language Model for fake-news detection & TEXT-RF, LR-Bias, Ternion, EANN, SpotFake, RumorLLM & \textcolor{blue}{English} & RumorLLM achieved F1-score: 0.8679, AUC-ROC: 0.9233 & Limited to plain text rumors, ethical concerns \\

\textcolor{blue}{2024} & Knowledge Enhanced Vision and Language Model for Multi-Modal Fake News Detection & IEEE TOM & Proposes knowledge-integrated model for fake news detection & NeuralTalk, att-RNN, MVAE, SpotFake, FND-SCTI, RIVF, MFN & \textcolor{blue}{English, Images} & Proposed model outperformed W/O KG, F1-score: 0.824 & Knowledge graph embedding updates impact performance \\
\bottomrule
\end{tabular}
\end{table*}

\renewcommand{\arraystretch}{1.2}
\begin{table}[]
    \caption{Urdu News Collection from various authentic sources.}
    \begin{center}
   
    \begin{tabular}{|p{0.3\textwidth}|p{0.04\textwidth}|p{0.03\textwidth}|p{0.03\textwidth}|}
    
        \hline
        \textbf{URL} & \textbf{Dataset count} & \textbf{Fake} & \textbf{Legit} \\
        \hline
        \url{https://huggingface.co/datasets/jawadhf/Labeled_urdu_news/blob/main/Labeled_Urdu_News.xlsx} & 4097 & 2455 & 1642 \\
        \hline
        \url{https://www.kaggle.com/datasets/tridata/notri-fact-real-and-unreal-urdu-news/data} & 13388 & 6677 & 6711 \\
        \hline
        \url{https://github.com/Sheetal83/Ax-to-Grind-Urdu-Dataset/blob/main/Combined%20.csv} & 10083 & 5053 & 5030 \\
        \hline
        \textbf{Total} & \textbf{27568} & \textbf{14185} & \textbf{13383} \\
        \hline
    \end{tabular}
    \label{tab1}
    \end{center}
\end{table}

Urdu Fake News Detection (FND) is addressed using an ensemble model based on transformer-based models. The dataset consisted of 1.08k fake and genuine news articles in Urdu. The ensemble model\cite{4} outperformed existing machine learning and deep learning models for the detection of Urdu fake news. Techniques such as model ensemble and the MV-V technique for stacking were employed to improve performance and avoid overfitting. A staking model combining three algorithms, including mBERT, XLNet, and XLM-RoBERTa, achieved impressive metrics: an F1-score of 0.924, an accuracy of 0.956, and an MCC value of 0.902.

The RumorLLM model \cite{3} improves fake news detection by leveraging rumor writing styles and content. It outperforms baselines in terms of F1 score and AUC-ROC. The model addresses category imbalance in datasets by working with diverse small-category samples. However, it is limited to plain text rumors and raises ethical concerns about potential misuse for spreading false information. A number of studies applied the pre-trained NLP model, ML, and DL model in detecting fake news\cite{13}\cite{24}\cite{25}\cite{26}.

The UQA dataset \cite{27}, generated from SQuAD2.0 via EATS, is crucial for Urdu QA. Evaluating Google Translator and Seamless M4T, the study benchmarks multilingual QA models (mBERT, XLM-RoBERTa, mT5) on UQA. Notably, XLM-RoBERTa-XL achieved 85.99 F1 score and 74.56 EM. The paper emphasizes NLP resource expansion to low-resource languages like Urdu \cite{28}, \cite{29}, demonstrating EATS' efficacy in dataset creation.

Researchers \cite{9} delve into machine learning for Urdu fake news detection, achieving high accuracy with ensemble models across nine domains. Challenges persist due to the language's low-resource nature and sparse multi-domain data availability. Their proposed classifier integrates features like TF-IDF, Bag of Words, stacking, and ensemble methods \cite{10a,10b}, yielding impressive metrics. 

Efforts to detect fake news and abusive language in Urdu, especially on social media\cite{11}, have seen advancements using machine learning models like SVM, Random Forest, CNN, and RNN. The CNN model notably achieved a remarkable 99\% accuracy in identifying fake news from Urdu tweets, while RNN and Random Forest also performed well, with accuracies around 91\%. However, challenges remain, including a lack of comprehensive research and limitations in existing datasets, suggesting the need for multilingual analysis and real-time monitoring to enhance detection systems' robustness.

This study proposes a large news dataset gathered from reliable World Wide Web (WWW) sources and normalizes and pre-processes it by combining many online resources, including research papers, GitHub, Hugging Face, and the famous Kaggle repository. The news is from the Asian region and covers numerous domains, including healthcare, politics, sports, entertainment, weather, education, etc., to increase the lexical diversity of the dataset. Journalists and experts wet the sample annotated dataset. This news comes from various famous and leading newspapers published daily online and presented in Table 1. Due to the exponential growth of the news, it is hard to find legit news, especially from social media news. So, we used the pre-trained models famous in Natural language processing (NLP), including classification and text generation, and fine-tuned those models on our proposed dataset and evaluated them by comparing them to the other ML and DL SOTA techniques. Table 1 demonstrates the news count and sources for our proposed dataset.

\begin{figure*}[]
 \centering
 \includegraphics[width=0.8\textwidth,height=11cm]
 {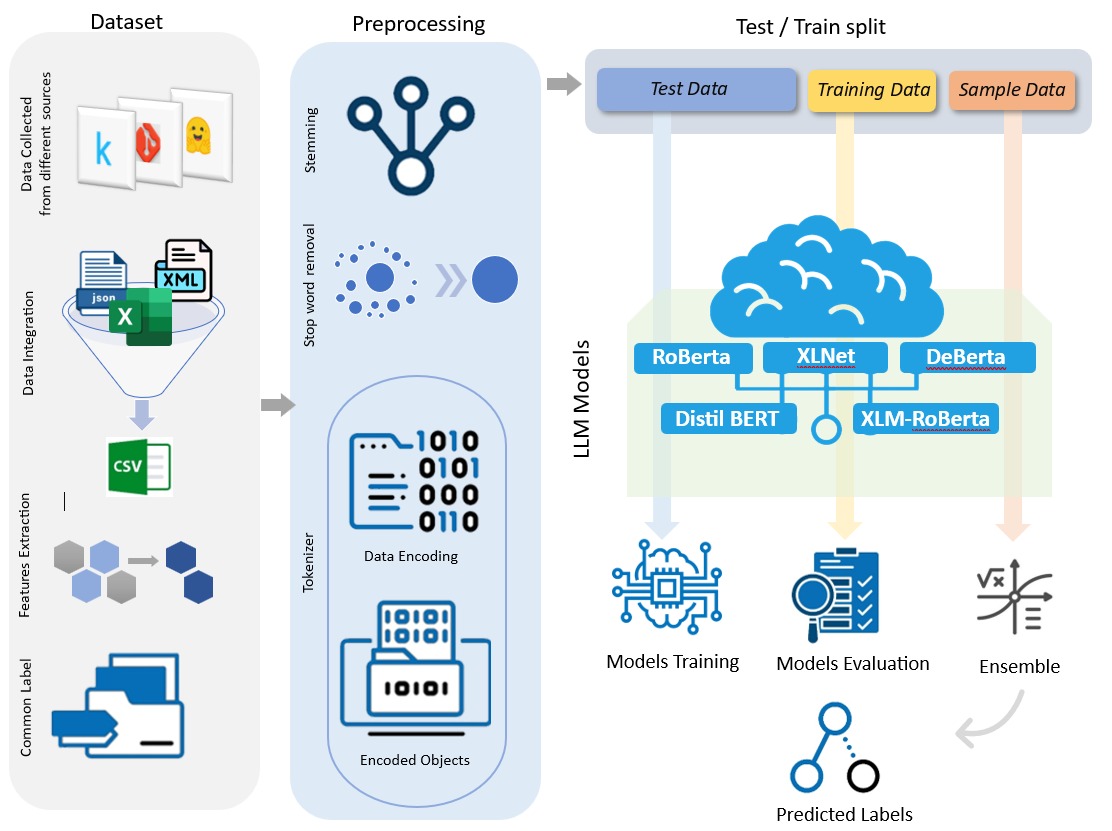}
 
\caption{ Our overall Proposed framework of ULLM on the LUND dataset}
\label{fig}
\end{figure*}

\section {Materials and Methods}

\subsection {Large Urdu News Dataset (LUND)}
Numerous benchmark datasets for Fake news detection have been developed in English, including famous LIAR, Fake, and legit datasets. Therefore, the data scientist works in the under-research area and develops a number of benchmark datasets in Urdu, considering the importance of benchmark corpus and algorithms that work on under-resourced languages. The Urdu dataset includes the UrduFake@FIRE2020\cite{2}, Ax-to-Grind Urdu\cite{4}, and Bil-FaND \cite{16} Urdu News \cite{9} datasets for fake news detection; however, some datasets are not publicly available, and some are not too large for the evaluation of large language models. Some are translated from English to Urdu. So, developing or fine-tuning LLM algorithms for under-resource languages like Urdu with low lexical diversity and words to evaluate and detect fake news is challenging. Therefore, it is necessary for an hour to introduce a comprehensive significant language dataset (LLD) for the Urdu language that is spoken around the globe. This research fused the first publicly available large datasets containing 27.5K news from authentic sources and developed the first large dataset to apply and fine-tune LLMs and evaluate AI and NLP algorithms for the specific language, Urdu shown in Table 2. The word cloud of legit and fake news is plotted in Figure 2.

\subsubsection{Public authentic repositories for data-gathering}\label{AA}
The availability of news datasets for detecting and mitigating fake news, particularly in the Urdu language, remains insufficient. However, a thorough review of research papers published by reputed publishers such as Elsevier and Springer, as well as top-tier conferences, provided access to some publicly available, authentic datasets. Despite this, the number of such datasets remains inadequate for constructing large-scale language datasets.
To address this gap, we explored the well-known data science community platform Kaggle, where we obtained fake news datasets after verifying their sources, including newspaper links and sampling methods. Each dataset was cross-validated against credible news sources to ensure authenticity. Additionally, continuous exploration of repositories such as GitHub and Hugging Face proved beneficial in retrieving further datasets. However, only those datasets with verifiable source origins were curated for inclusion in our study. Figure 3 illustrates the detailed proposed framework of ULLM on the LUND dataset, demonstrating the integration of these curated datasets into our approach.

\subsection{Pre-trained Large Language Transformer-based Models (LLTMs)}
\subsubsection{Problems faced in data gathering}
The requirement for fake news detection is to gather both fake and legitimate news datasets from authentic sources. For this purpose, the exploration is not just limited to research articles but also other famous community sources, making the holistic dataset-gathering process so challenging.
\subsubsection{Inconsistency of data formats}
As data is gathered from many sources, different sources use different formats (e.g., CSV, JSON, XML), making merging data into one difficult. Merging data from different resources also creates duplication and incompleteness, which has resulted in an inconsistent dataset.

\subsubsection{Data Inconsistency}
Different Definitions of Fake News: Various datasets have defined fake news differently, including misinformation, disinformation, Fake or real, legit, satire, or biased news. So, we standardized these definitions, which are crucial but challenging.

\subsubsection{Labeling Inconsistencies}
Datasets have used different labelling schemes (e.g., true/false, legit/fake, real/fake, or various categories of misinformation), leading to inconsistency. FND is a binary class problem, so we transform True, Partly True, and Half True labels into fake news. However, we resolved this into binary classification of legit or fake news.
\subsubsection{Domain and region differences}
The datasets have multi-domain and cross-domain data, so as a result, a domain like politics, cricket, sports, economy, technology, weather, etc., has a majority of the news compared to women’s rights, religion, education, and health, which has limited news data. So we tackle the imbalance domain data issue keeping in view that the records are balanced to overcome domain biases, and as a result, we get lexical diversity in our dataset.
\subsubsection{Class imbalance}
The dataset we gathered from various resources has imbalanced class data, which affects the algorithm's performance and causes overfitting. The datasets, when combined, highly imbalanced data as legit news has much more data than fake news, and there isn’t any prominent fact checker website in the Urdu language to check its authenticity. That is why we picked datasets from fake news from Kaggle, Git Hub, and Hugging Face for standard distribution; otherwise, we need to apply for stratified sampling on imbalanced news items.

\subsubsection{Data Duplication }
Some attributes are discarded because they are present in some datasets but not in other datasets. We carefully chose the common attributes in every dataset, keeping synonyms and homonyms in mind, and gave each attribute a standard name that is present in the majority. We also removed the duplicate entries during the merging of the dataset to avoid skewed results. Slightly altered news was also present in the merged dataset, so deduplication was hard, but we removed that news as well.  

\subsection{Dataset statistics}

The overall dataset contains 67.91k words in both fake and legitimate Urdu news text. Fake News (FN) and Legit News (LG) contain 27.10k common words. 80\% of the dataset is used to train the LLM algorithm, while 20\% is used to validate and test the developed algorithm. Table 3 and Figure 4 show the dataset’s statistics with sample examples.

\begin{table}[]
    \centering
    \caption{Illustrate the division for training and testing samples }
    
    \begin{tabular}{|c|c|c|c|c|}
        \hline
        \textbf{Total  News  } & \textbf{Legit } & \textbf{Fake } & \textbf{Training  } & \textbf{Testing  } \\
        \hline
        27568 & 13383 & 14185 & 22055 & 5513 \\
        \hline
        
    \end{tabular}
    \label{tab:sample-distribution}
\end{table}

 \begin{figure}[!t]
    \centering
    \includegraphics[width=0.9\linewidth]{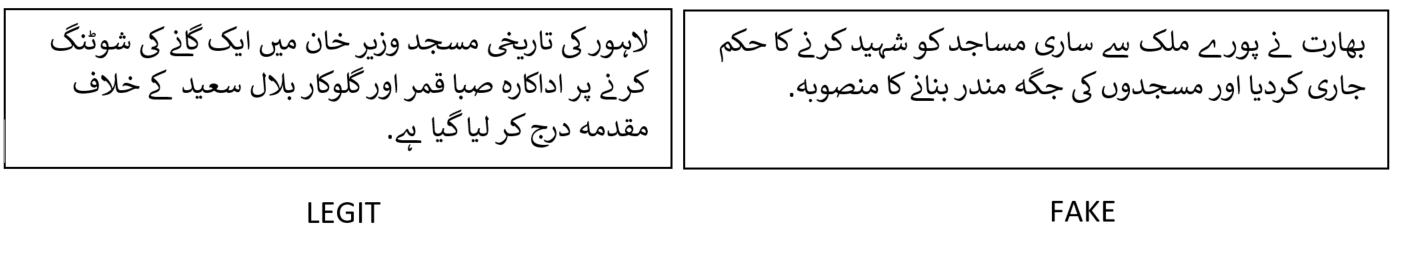}
    \captionof{figure}{Sample data of fake and legit Urdu News}
    \label{fig:enter-label}
    \end{figure}

\subsection{Urdu news Dataset(LUND) Pre-Processing }
Raw data is required to be cleaned and pre-processed before being fed into the models for training because it may affect the training time and classification of algorithms. Therefore, IP addresses, URLs, special characters, and non-alphanumeric characters should be removed. The following sort of action is taken on the raw dataset.

\subsubsection{Sentence segmentation and stemming, removing stop words }

Sentence segmentation is used to process the dataset's input string into relevant sentences. For information, the full stop (.) is added at the end of the simple sentence. For clarity, the (,) is added between the sentences, and a question mark (?) is added to indicate the question. Stemming is used to increase the clarity of the word and minimize the confusion between the same words. So, by using the stemming algorithm, we transform the words with suffixes to their base form. Figure 5 showcases the techniques used in tokenizer and encoded objects.
%
%
Removing stop words is essential in the Urdu language because of limited semantic and contextual value. 
It is also considered a noise-removal technique. 
For clarity, some example figures are added from our proposed methodology dataset.

\begin{figure}[!t]
 \centering
 \includegraphics[width=0.2\textwidth,height=3cm]
 {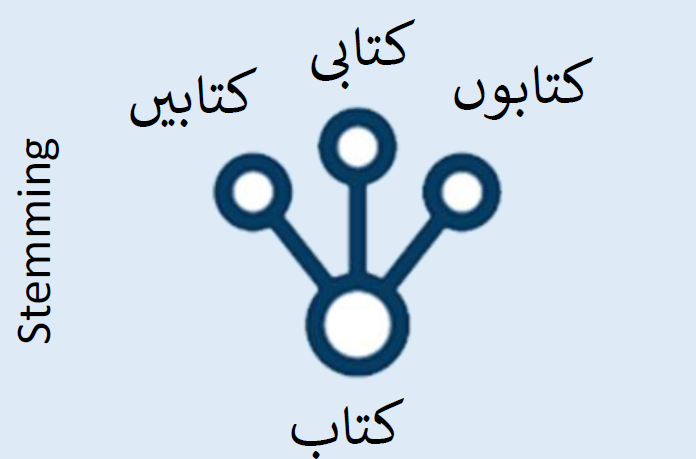}
 \includegraphics[width=0.2\textwidth,height=3cm]
 {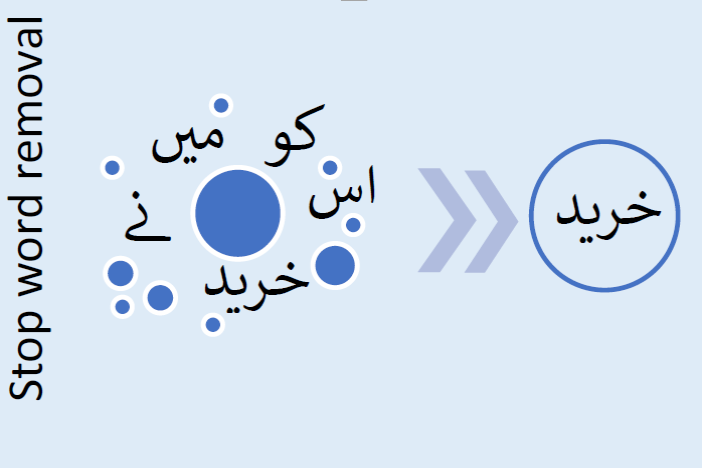}
\caption{Stemming and stop word removal technique}
\label{fig}
\end{figure}
\subsubsection{Tokenization with encoding objects}

News words in a sentence are tokenized into each item for interpretation of our proposed Urdu language NLP models. These tokens are transformed into various codes in the data encoding process and then converted into encoded objects. Figure 6 showcases the techniques used in tokenizer and encoded objects.

\begin{figure}[!t]
 \centering
 \includegraphics[width=0.2\textwidth,height=3cm]
 {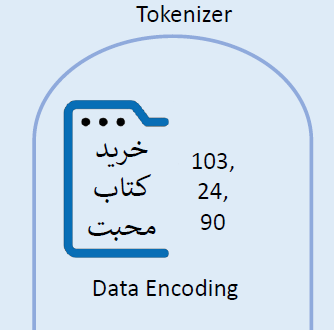}
 \includegraphics[width=0.2\textwidth,height=3cm]
 {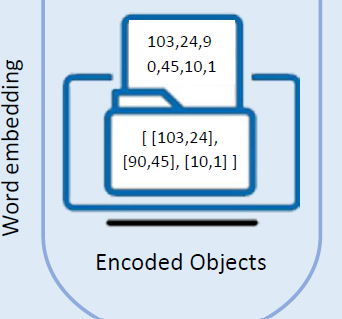}
\caption{Demonstration of tokenizer and encoded objects}
\label{fig}
\end{figure}

\subsubsection{Lexical and semantic Feature Extraction}
Our methodology utilizes various features to improve AI model performance in fake news detection. Frequency measures how often a word appears in a news item, indicating its importance, while document frequency reveals its prevalence across the dataset. This affects the TF-IDF score, which rises with word repetition in a document and falls as more documents contain the word. We extract lexical, sentimental, and n-gram features for machine learning algorithms. Additionally, we use BERT embedding with deep learning algorithms and applied SOTA NLP pre-trained models to capture contextual information and enhance model robustness. 
\begin{figure*}[]
\centerline{\includegraphics[width=0.8\textwidth,height=6cm]{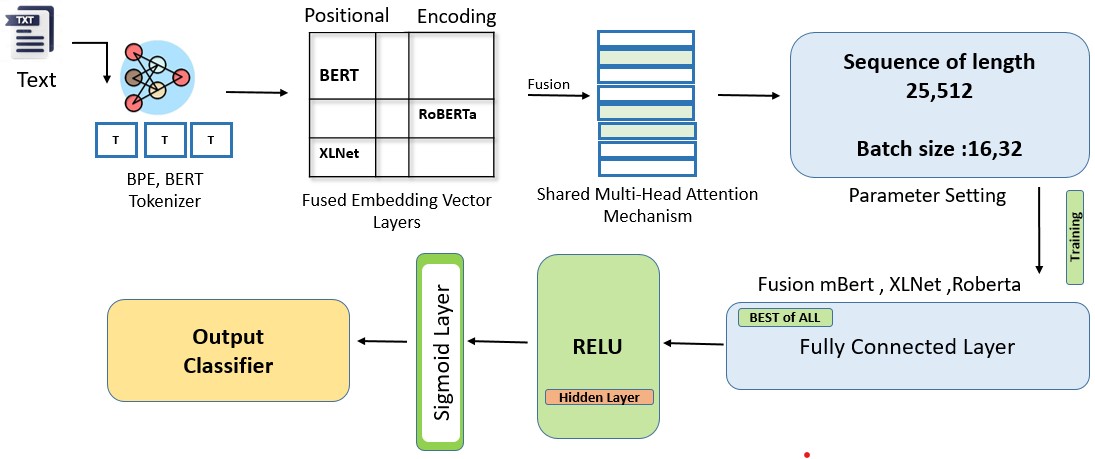}}
\caption{Demonstrates our proposed unified model.}
\label{fig}
\end{figure*}
\subsection{Pre-trained Large Language Transformer-based Models (LLTMs)}

Large Language Models (LLMs) have transformed NLP in recent years. They enable Transfer Learning (TL), where you can use your attained knowledge during pre-training to leverage for numerous tasks, including sentiment analysis\cite{26}, questioning and answering \cite{27}, translation and text generation\cite{12}, and summarization\cite{13, 13a}. Numerous recent feature engineering techniques like lexical, sentimental, and N-gram with certain activation functions and pooling strategies aid the LLM models of the NLP algorithm to reach new heights. We have chosen five LLM models from the Hugging Face website that are already trained on massive multilingual corpora, including Wikipedia, Blogs, Academic Journals, etc. These pre-trained models, including XLNet, mBERT, XLM-Roberta, DistilBert, Roberta, and DeBerta,\cite{20}, \cite{22}, \cite{25}, \cite{27} are used in our research studies to benchmark the dataset. 
DistilBERT is a version of Bidirectional Encoder Representations from Transformers (BERT) developed by Hugging Face. It aims to maintain robust performance while being lightweight and using distillation knowledge during pre-training. However, XLNet is the generalized Auto-Regressive (AR) model used in compressive language understanding. The representation shows that its values rely on the previous ones because it works in a sequence. The autoregressive (AR) model of order $p$ is defined as:

\begin{equation}
    \text{AR}(p) = \left[ X_t = c + \sum_{i=1}^{p} \phi_i X_{t-i} + \varepsilon_t \right]
\end{equation}
Where $X_t$ represents the value at time $t$. $c$ is a constant term. $\phi_i$ are the autoregressive coefficients for lags $1$ to $p$. $X_{t-i}$ refers to the value at time $t-i$. $\varepsilon_t$ is the error term (usually assumed to be normally distributed with mean $0$). Due to its AR nature, it overcomes the limitations of BERT.

XLM-RoBERTa is a famous low-resource language model. It's a large transformer-based model trained on 2.5TB of web data. This model achieves state-of-the-art (SOTA) results and outperforms many multilingual and cross-lingual corpora in $100$ languages. mBERT is a strong NLP masked language model published by Google. It is also trained on multilingual and cross-lingual large corpora. The Masked Language Model (MLM) objective function is defined as:

\begin{equation}
    \text{MLM} = \sum_{i=1}^{N} \log p(x_i | x_{<i}, x_{>i})
\end{equation}

$N$ is the total number of tokens in the sequence. $x_i$ represents the $i$-th token. $x_{<i}$ and $x_{>i}$ denote the tokens before and after the $i$-th token, respectively. 

The Masked Language Model (MLM) generally works on three core steps: Masking, prediction, and Training objectives to maximize the likelihood of correct masked words given the context.

DeBERTa is a self-supervised learning model pre-trained on a large corpus with an enhanced decoding mechanism. Its aim is universal natural language understanding and representation.

\subsection{Unified Large Language Model (ULLM)}
This section discusses how multiple pre-trained large language models (LLMs) were used and combined to improve performance on a given task, following a method called stacking.
\\
\subsubsection{Model Selection and Fusion}

Several different LLMs, including GPT-2, XLNet, RoBERTa, XLM-RoBERTa, DistilBERT, mBERT, and DeBERTa, were selected. These models were initially trained on large multilingual corpora, meaning they can understand multiple languages \cite{28}.

The idea is to leverage the strengths of each model by combining their predictions to form a unified, more powerful model. This is called stacking, a machine learning technique where multiple models’ predictions are merged to improve overall performance \cite{29}. The goal is for the combined model to be more robust and accurate than any individual model alone, as shown in figure 7.
\\\begin{table*}[htbp]
    \centering
    \caption{Performance comparison of large language models on the developed dataset.}
    \small 
    \begin{tabular}{|p{2.2cm}|p{1.6cm}|p{1.8cm}|p{1.6cm}|p{1.6cm}|p{1.6cm}|p{1.6cm}|p{1.6cm}|}
        \hline
        \textbf{Parameter} & \textbf{XLNet} & \textbf{XLM-RoBERTa} & \textbf{mBERT} & \textbf{DistilBert} & \textbf{Roberta} & \textbf{DeBerta} & \textbf{Ours} \\
        \hline
        Accuracy & 0.855 & 0.860 & 0.830 & 0.845 & 0.815 & 0.820 & 0.959 \\
        \hline
        Precision & 0.917 & 0.928 & 0.845 & 0.882 & 0.812 & 0.892 & 0.961 \\
        \hline
        Recall & 0.889 & 0.922 & 0.836 & 0.877 & 0.868 & 0.831 & 0.958 \\
        \hline
        F1-Score & 0.903 & 0.925 & 0.841 & 0.879 & 0.839 & 0.861 & 0.960 \\
        \hline
    \end{tabular}
\end{table*}

\begin{table*}[htbp]
    \centering
    \caption{Performance comparison of different models on the developed dataset.}
    \small 
    \begin{tabular}{|p{1.4cm}|p{1.3cm}|p{1.2cm}|p{1.2cm}|p{1.2cm}|p{1.2cm}|p{1.2cm}|p{1.2cm}|p{1.2cm}|p{1.2cm}|p{1.2cm}|}
        \hline
        \textbf{Model} & \textbf{KNN} & \textbf{SVM} & \textbf{DT} & \textbf{RF} & \textbf{LR} & \textbf{NB} & \textbf{GB} & \textbf{MLP} & \textbf{CNN} & \textbf{LSTM} \\
        \hline
        Accuracy & 0.904 & 0.834 & 0.826 & 0.953 & 0.815 & 0.812 & 0.751 & 0.958 & 0.932 & 0.784 \\
        \hline
        F1-Score & 0.861 & 0.856 & 0.827 & 0.936 & 0.806 & 0.770 & 0.768 & 0.958 & 0.979 & 0.818 \\
        \hline
        Precision & 0.916 & 0.895 & 0.815 & 0.916 & 0.811 & 0.819 & 0.753 & 0.945 & 0.971 & 0.822 \\
        \hline
        Recall & 0.814 & 0.821 & 0.839 & 0.956 & 0.801 & 0.727 & 0.783 & 0.972 & 0.987 & 0.814 \\
        \hline
    \end{tabular}
\end{table*}

\subsubsection{Model Tokenizers and Embedding Techniques}

Tokenizers such as Byte Pair Encoding (BPE), BertTokenizer, and XLMRobertaTokenizer were used with different models, which are responsible for breaking down text into smaller units (tokens). These tokenizers match the training approach of each model (BPE for GPT-2, BertTokenizer for BERT-based models, and XLMRobertaTokenizer for XLM-RoBERTa).
Embedding techniques such as those from BERT and XLNet were employed to map text data into high-dimensional vectors for efficient model training.
\\
\subsubsection{Hyperparameter Tuning}

Hyperparameters (such as batch size, epochs, and learning rate) were fine-tuned to achieve optimal results \cite{30}. A batch size of 16-32 and 10 training epochs were used to train the models. The learning rate was set to 
$4 \times 10^{-10}$ , and optimizers like AdamW and SGD were applied for updating model weights during training.
 \\
\subsubsection{Model Training}

Each model was trained separately on the given datasets, and the trained models were then tested on sample datasets. The models predicted the class of the input text (e.g., whether it was "legit" or "fake").
\\
\subsubsection{Prediction Aggregation}

The predictions from all the models were gathered. Since each model might have a slightly different prediction, the final decision is made by majority voting, where the most frequently predicted class across all models is chosen.
\\
\subsubsection{Performance Metrics}

The performance of the unified model was evaluated using common metrics like F1 score, accuracy, precision, and recall. These metrics help to assess how well the model is performing in terms of classification (e.g., distinguishing fake vs. legit content) \cite{31},\cite{32}.
\section{Results and Discussion}
The performance of the models is evaluated using the confusion matrix and calculating their precision and recall. XLNet achieves 0.855 accuracy and 0.903 f1 measure. XLM-RoBERTa and DistilBert determine 0.860, 0.845 accuracy and 0.925, 0.79 f1 score. Meanwhile, mBERT shows 0.830 accuracy with an f1 score of 0.903. DeBerta demonstrates accuracy and f1 scores of 0.820 and 0.861, respectively. Our proposed ULLM technique resultantly achieved optimal results with accuracy of 0.959, precision and recall of 0.961, 0.958, and an F1 score of 0.960. Table IV and Figure 8 demonstrate the comparison of all the LLMs with our unified model.

\begin{figure}[]
\centerline{\includegraphics[width=1\linewidth]{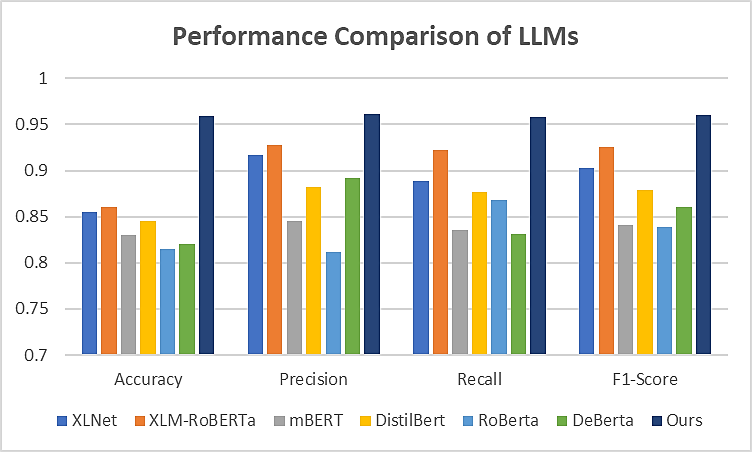}}
\caption{Demonstrates
the comparison of all the LLMs with our unified model}
\label{fig}
\end{figure}

\begin{figure}[]
\centerline{\includegraphics[width=1\linewidth]{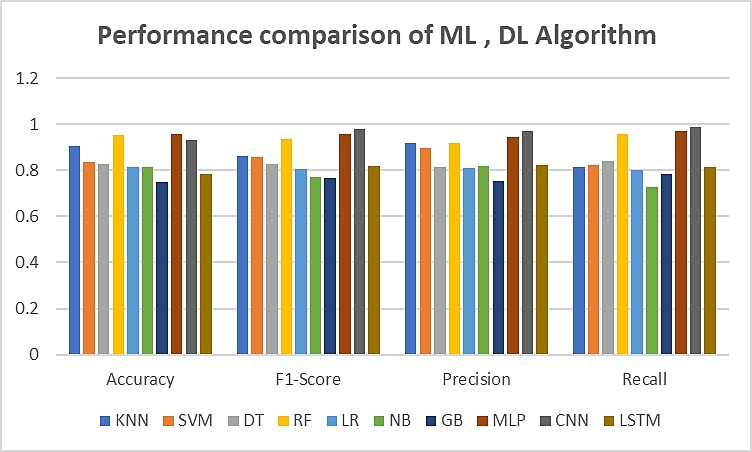}}
\caption{llustrates the comparison of all ML and DL
model.}
\label{fig}
\end{figure}

Performance evaluation was conducted using state-of-the-art machine learning (ML) and deep learning (DL) algorithms, as discussed in the literature review, on our benchmark dataset. The results indicate that the Multilayer Perceptron (MLP) outperforms traditional ML classifiers, including K-Nearest Neighbors (KNN), Support Vector Machine (SVM), Decision Tree (DT), Random Forest (RF), Logistic Regression (LR), Naïve Bayes (NB), and Gradient Boost (GB)\cite{24}.

Among deep learning models, Convolutional Neural Networks (CNNs) performed better than Long Short-Term Memory (LSTM) networks, highlighting CNN’s effectiveness in feature extraction for text classification\cite{25}.

Our proposed dataset is comparable in terms of reliability and achieves better diversity and generalization compared to previously available datasets. To enhance model robustness, we trained our model on a fusion of multiple existing datasets and performed cross-dataset evaluation, assessing its generalization ability across different data distributions. The results confirm that the proposed approach improves model robustness and adaptability.

Furthermore, the findings demonstrate that pre-trained language models consistently outperform traditional ML and DL models on our proposed integrated benchmark dataset. However, our unified technique, which integrates multiple pre-trained models through stacking, achieves superior performance compared to individual pre-trained models on the Urdu Large Language Dataset (ULLD). A comprehensive comparison of all ML and DL models, along with their performance metrics, is provided in Table V and Figure 9.

\section{Conclusions and future work}

This study aims to address multiple issues in the literature regarding fake news detection in the Urdu language. The key findings include the curation and evaluation of the first comprehensive large Urdu language dataset, comprising 27,410 instances of fake and legitimate news across 23 diverse domains. Our proposed unified model, leveraging the strengths of six well-known NLP LLM models previously trained on large multilingual datasets with different  fused embedding, share attention,  fine-tuning, and activation functions, outperformed all previously presented ML and DL models in the literature. The achieved accuracy and F1 scores of 0.959 and 0.960, respectively, demonstrate the unified model's effectiveness in benchmarking large new datasets. Human intelligence vetted the sample dataset both before and after applying artificial intelligence techniques, enhancing the dataset's quality and the proposed model's outcomes.

In the future, the developed dataset can be used with other embedding and ensemble models to achieve improved results. The proposed unified model can be applied to multilingual datasets to test its efficacy. Additionally, the dataset can be evaluated using other performance metrics to verify its validity and authenticity. A fact-checking algorithm can be developed based on the proposed dataset to assess news biases. Misclassified samples can be investigated through human expert judgment. Finally, research can focus on creating a smaller language model for accurate news classification for others task with improved time complexity.

\section*{Declaration of Interest}
The authors declare that they have no known competing financial interests or personal relationships that could have appeared to influence the work reported in this document.

\section*{Data availability}
The data source will be provided on demand.
\section*{Acknowledgement}
The authors extend their appreciation to the Deanship of Research and Graduate Studies at King Khalid University for funding this work through Small Group (Reseach Project under grant number RGP1/243/45).

\vspace{12pt}
\color{red}

\end{document}